\documentclass{article}

\usepackage{microtype}
\usepackage{graphicx}
\usepackage{subfigure}
\usepackage{booktabs} 
\usepackage{etoolbox}

\AfterEndEnvironment{figure}{\vskip-1ex}

\usepackage{hyperref}

\graphicspath{ {./Images/} }
\usepackage[accepted]{icml2020}

\icmltitlerunning{Reinforcement Learning Generalization with Surprise Minimization}

\begin{document}
\twocolumn[
\icmltitle{Reinforcement Learning Generalization with Surprise Minimization}




\icmlsetsymbol{equal}{*}

\begin{icmlauthorlist}

\icmlauthor{Jerry Zikun Chen}{uoft}

\end{icmlauthorlist}

\icmlaffiliation{uoft}{Department of Computer Science, University of Toronto, Toronto, Canada}

\icmlcorrespondingauthor{Jerry Zikun Chen}{jzchen@cs.toronto.edu}

\icmlkeywords{Machine Learning, Reinforcement Learning, ICML}

\vskip 0.3in
]

\printAffiliationsAndNotice{}  

\begin{abstract}
    Generalization remains a challenging problem for deep reinforcement learning algorithms, which are often trained and tested on the same set of deterministic game environments. When test environments are unseen and perturbed but the nature of the task remains the same, generalization gaps can arise. In this work, we propose and evaluate a surprise minimizing agent on a generalization benchmark to show an additional reward learned from a simple density model can show robustness in procedurally generated game environments that provide constant source of entropy and stochasticity.
    
\end{abstract}

\section{Introduction}
Reinforcement learning (RL) algorithms hold great promises in the pursuit of artificial intelligence as demonstrated by beating human players in board games \cite{Go, Chess}, reaching expert level in video games \cite{Starcraft, Dota}, and improving complex robotic control tasks \cite{sac}. However, deep RL agents are often hampered by their struggle to generalize in new environments, even when the semantic nature of the task remain similar \cite{gen_dqn, zhang_b, image}. It is evidenced that they are prone to overfitting by memorizing experiences from the training set \cite{zhang_b}. Therefore, a near-optimal policy can nonetheless produce generalization gaps in unseen game levels \cite{CoinRun, Procgen}. This problem makes RL agent unreliable for real world applications where robustness is important.

Recent work (SMiRL) \cite{SMiRL} inspired by the free energy principle \cite{FEP} has shown that intelligent behaviours can emerge from minimizing expected future surprises in order to maintain homeostasis. This is motivated by the entropy-increasing nature of the real world that can provide constant novelty without the need for explicit intrinsic motivation to explore. A SMiRL agent alternates between learning a density model that estimates the marginal state distribution under its policy, and improving the policy to seek more predictable stimuli. Probability of an observed state is optimized and predicted by decoding the maintained density model learned from history. Theoretically, this objective of minimizing future surprises in entropic environments tightens a lower bound on the entropy of the marginal state distribution under the policy. Adding a surprise minimizing reward to the raw episodic task reward also shows faster reward acquisition. This framework can potentially help the agent achieve better generalization since it is integral that a surprise minimizing agent learns a robust policy in order to perform well in a dynamically changing environment. When the agent minimizes future surprises throughout training, a bias toward stability and predictability can be learned to counteract the prevailing source of entropy from the environment, which should be present in the test set as well. 

To evaluate generalization performance on diverse, and randomized training and test environments, a good candidate is the Procgen benchmark \cite{Procgen}. It consists of 16 game environments designed using procedural content generation \cite{procedural}. Compared to human-generated level layout and progression for each game that are commonly seen in RL game benchmarks, procedural generation creates highly randomized content for each episode that is suitable for evaluating generalization. To do well on the test set, an agent needs to be encouraged to learn a policy that is robust against changes in variables such as level layout, initial location of entities, background designs, et cetera. We show that adding surprise minimizing information through rewards learned by the density model on states of the training history can improve generalization on procedurally generated random game levels that are unseen during training. Since the history consists of episodes in levels that are dynamic and unpredictable, to do well, the agent must learn to stabilize amidst the changing nature of its environment and acquire robust skills through minimizing the chances of seeing surprising states.

\section{Related Work}
RL benchmarks dealing with generalization have emerged in recent years. Sticky actions \cite{sticky_action} and random starts \cite{random_start} are methods to add stochasticity to the Atari benchmarks. Procedural generation is used to create diverse game environments \cite{video_game, procgen2}. In procedurally generated Obstacle Tower \cite{obstacle_tower}, the agent is required to learn both low-level control and high-level planning problems. Generalization performance of baseline algorithms like PPO \cite{PPO} and Rainbow \cite{rainbow} are poor on the Sonic benchmark \cite{sonic}, which is targeted at few-shot learning. Safety Gym \cite{safety_gym} consists of a suite of 18 high-dimensional continuous control environments for safe exploration under the constrained MDP framework. While their environments do have a high level of randomization, their main objective is to enforce safety and to minimize the constraint cost. Various techniques have been proposed to achieve better generalization for RL agents. It has been shown \cite{Transfer} that robust representations can be learned by the RL agent if diverse observations are provided during training. One example of this involves adding a convolutional layer in between the input image and the neural network policy \cite{data_aug}. This layer is randomly initialized at every iteration to create a variety of randomized training data. \citealt{zhang_b} report insightful discussions on the nature of RL over-fitting. They experiment on procedurally generated gridworld mazes and find that agents have a tendency to memorize specific levels in a given training set and have risk of overfitting robustly. Experiments on the CoinRun game \cite{CoinRun} investigates the impact of supervised learning regularization techniques including L2 regularization, dropout, data augmentation, and batch normalization, which all help narrow the generalization gap. Similar to \citealt{zhang_b}, they also found that injecting stochasticity with $\epsilon$-greedy action selection and increasing entropy bonus in PPO can both improve generalization. Importantly, stochasticity helped generalization to a greater extent than regularization techniques from supervised learning. However, it is believed that forced randomness that is not sourced from the environment can have a negative impact on the learning agent \cite{random_start}. Rather than altering the action selection procedure, we inject stochasticity through a different channel in the reward by learning state distributions from highly stochastic and dynamic environments in the Procgen benchmark. Since observed states are hard to predict with high entropy in regards to the state marginal distribution, the source of the stochasticity lies inherently from the environment itself. 

\section{Methodology}
In \citealt{Procgen}, agents are trained using PPO \cite{PPO} for 200M time steps under the hard difficulty. Under this setting, training requires approximately 24 GPU-hours per environment. In our case, for computational efficiency on a single GPU on Colab, we follow the suggestion from \citealt{Procgen}, both of our training and testing are done under the easy difficulty setting for 25M steps. A finite set of levels is use as the training set and the full distribution of levels are in the test set to evaluate generalization. Similar to \citealt{Procgen}, we use OpenAI baselines \cite{baselines} implementation of PPO, which has an IMPALA \cite{impala} network as the policy. For clarity, all curves in Section 4 are exponentially smoothed with weight 0.5.

\begin{figure}[ht]
\begin{center}
\centerline{\includegraphics[width=7cm]{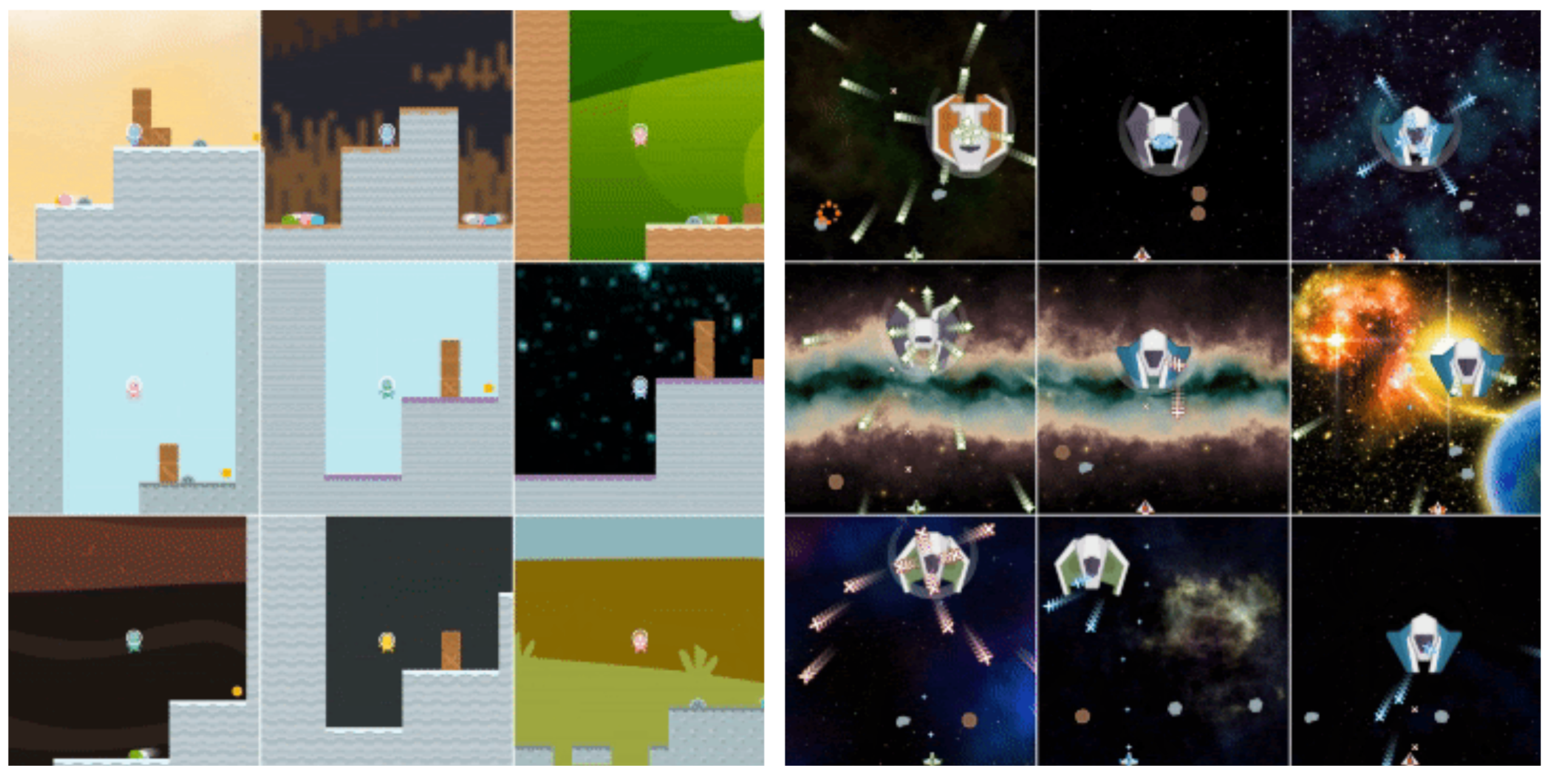}}
\caption{CoinRun and BossFight}
\label{game_sample}
\end{center}
\end{figure}

We train and evaluate the PPO baseline on the game CoinRun \cite{CoinRun}, the inaugural environment in the Procgen benchmark, used by other works. For further comparison, we also run experiments on another game BossFight, which has a more static observation background (see Figure~\ref{game_sample}).

Following \citealt{SMiRL}, we implement the surprise minimizing reward $r_{SM}$ based on a representation density model $p_{\theta_t}(s)$ as shown in Algorithm~\ref{alg:1}. This is an estimate of the true marginal state distribution under the policy. This model is used to construct the surprise minimizing reward. In the surprise minimization framework, one may think that SMiRL will seek degenerate behaviours that optimize for more familiar states that are high in $p_{\theta_t}(s)$ at the current step. But similar to an agent optimizing for sparse future rewards that might only appear at the end of the episode, the optimal policy does not simply visit states that have high probabilities now, but also those that can update $p_{\theta_t}(s)$ such that it provides high likelihood to the states that it sees in the long term.

\subsection{Normal Distribution} In Algorithm \ref{alg:1}, we first model $p_{\theta_t}(s)$ using independent Gaussian distributions for each dimension in our observations. A buffer is created for storing the most recent batches in the history during training. The size of the buffer is 20 times the mini-batch size. Observations in each batch $s_t$ contains 16384 RGB frame observations with dimensions 64 $\times$ 64 $\times$ 3. These observations may consist of multiple episodes depending on the agent's performance. We transform and normalize all RGB observations to 64 $\times$ 64 gray-scale images before they are stored in the buffer. Before each mini-batch update to our policy, a surprise minimizing reward is computed from the buffer. Given a particular state $s$ in the buffer $D$, the SM reward is computed as:

$$r_{SM}(s_t) = - \sum_{i}(\log\sigma_{i} + \frac{(s_i - \mu_i)^2}{2\sigma_i ^2})$$

where $\mu_i$ and $\sigma_i$ are the sample mean and standard deviation of the $i^{th}$ dimension calculated across each state in the data buffer. $s_i$ is value of the $i^{th}$ dimension in $s$. This reward is computed for each batched observations, rather than for individual observations in each episode in SMiRL \cite{SMiRL}. Therefore, this can be seen as an instance of non-episodic surprise minimization. This is also the case for the VAE method in Algorithm \ref{alg:2}.

\begin{algorithm}[tb]
    \caption{PPO + Normal Algorithm}
    \label{alg:1}
\begin{algorithmic}[1]
    \STATE Initialize game environment $Env$
    \STATE Initialize IMPALA policy $\pi_{\phi_0}$
    \STATE Initialize Data Buffer $D_0$
    \FOR {$t = 1, \cdots, T$} \do \\
        \STATE \# collect a batched experiences from current policy
        \STATE \# each batch $\mathbf{b}_t$ consists of multiple episodes
        \STATE $\mathbf{b}_t = \{\mathbf{s}_t, \mathbf{a}_t, \mathbf{r}_t\} \leftarrow Env(\pi_{\phi_{t-1}})$ 
        \STATE $D_{t} \leftarrow D_{t-1} \cup \{greyscale(\mathbf{s}_t)\}$ 
        \STATE \# computed from $D_{t}$ for 
        \STATE \# each dimensions across states
        \STATE $\theta_t = \{\mu_t, \sigma_t\} \leftarrow D_t$ 
        \STATE \# Normal SM rewards for each state in $\mathbf{s}_t$
        \STATE $\mathbf{r}^t_{SM} = \log p_{\theta_t}(\mathbf{s}_t)$  
        \STATE $\phi \leftarrow PPO(\phi_t,  \{ \mathbf{s}_t, \mathbf{a}_t, \mathbf{r}_t + \alpha \mathbf{r}^t_{SM} \})$
    \ENDFOR
\end{algorithmic}
\end{algorithm}

\subsection{Variational Autoencoder} Additionally, as presented in Algorithm \ref{alg:2}, we take a variational autoencoder (VAE) instead of Normal density to learn a richer representation from observations from batches. We trained the VAE with raw RGB observations without adjustment. Following \citealt{SMiRL}, instead of a data buffer, we train this VAE online across all episodes during training since it requires more data. Distinct from the VAE prior $p(\mathbf{s})$, a batch-specific distribution $p_{\theta_t}(\textbf{z})$ is tracked to compute the SM reward. $p_{\theta_t}(\textbf{z})$ is represented as a normal distribution with diagonal covariance. Parameters of this distribution are computed from the VAE encoder outputs of the observations in batch $\mathbf{s}_t$ with size $B$ (line 9 in Algorithm~\ref{alg:2}):

$$z_j = \mathrm{E}[q(z_j|s_j)], \forall s_j \in \mathbf{s}_t, j \in \{1, \cdots, B\}$$
$$\mathbf{\mu} = \frac{\sum_{j=0}^B z_j}{B+1}, \mathbf{\sigma} = \frac{\sum_{j=0}^B (\mu - z_j)^2}{B+1}, \theta_t = \{\mathbf{\mu}, \mathbf{\sigma}\}$$

To compute the SM rewards for each observation, we take the log probability $\log p_{\theta_t}(z_j)$ of this normal distribution evaluated at each $z_j$ in the batch. 

In both Normal and VAE cases, the SM reward is added to the raw episodic reward for the specific task via the equation: $$r_{combined}(s) = r_{task}(s) + \alpha r_{SM}(s)$$ $\alpha$ is a hyper-parameter selected to balance the magnitudes of the two rewards. Generalization in the additional SM reward settings are compared against the PPO baselines. The evaluation score is the mean raw task reward across all episodes for each mini batch update achieved during training and testing. It ranges from 0 to 10 for CoinRun and 0 to 12 for BossFight.

\begin{algorithm}[tb]
    \caption{PPO + VAE Algorithm}
    \label{alg:2}
\begin{algorithmic}[1]
    \STATE Initialize game environment $Env$
    \STATE Initialize IMPALA policy $\pi_{\phi_0}$
    \STATE Initialize VAE$_{\psi_0}$
    \FOR {$t = 1, \cdots, T$} \do \\
        \STATE $\mathbf{b}_t = \{\mathbf{s}_t, \mathbf{a}_t, \mathbf{r}_t\} \leftarrow Env(\pi_{\phi_{t-1}})$
        \STATE \# update VAE parameters
        \STATE $\psi_t \leftarrow SGD(\{\mathbf{s}_t\})$ 
        \STATE \# produce latent representation
        \STATE $\mathbf{z}_t \leftarrow$ VAE$_{\psi_t}(\{\mathbf{s}_t\})$ 
        \STATE \# compute parameter of diagonal  Gaussian from $\mathbf{z}_t$ 
        \STATE $\theta_t = \{\mu_t, \sigma_t\} \leftarrow \mathbf{z}_t$ 
        \STATE $\mathbf{r}^t_{SM}$ = $\log p_{\theta_t}(\mathbf{z}_t)$ \# VAE SM rewards
        \STATE $\phi \leftarrow PPO(\phi, \{ \mathbf{s}_t, \mathbf{a}_t, \mathbf{r}_t + \alpha \mathbf{r}^t_{SM} \})$
    \ENDFOR
\end{algorithmic}
\end{algorithm}

\section{Experimental Results}

\subsection{Normal Distribution}
\begin{figure}[ht]
\begin{center}
\centerline{\includegraphics[width=\columnwidth]{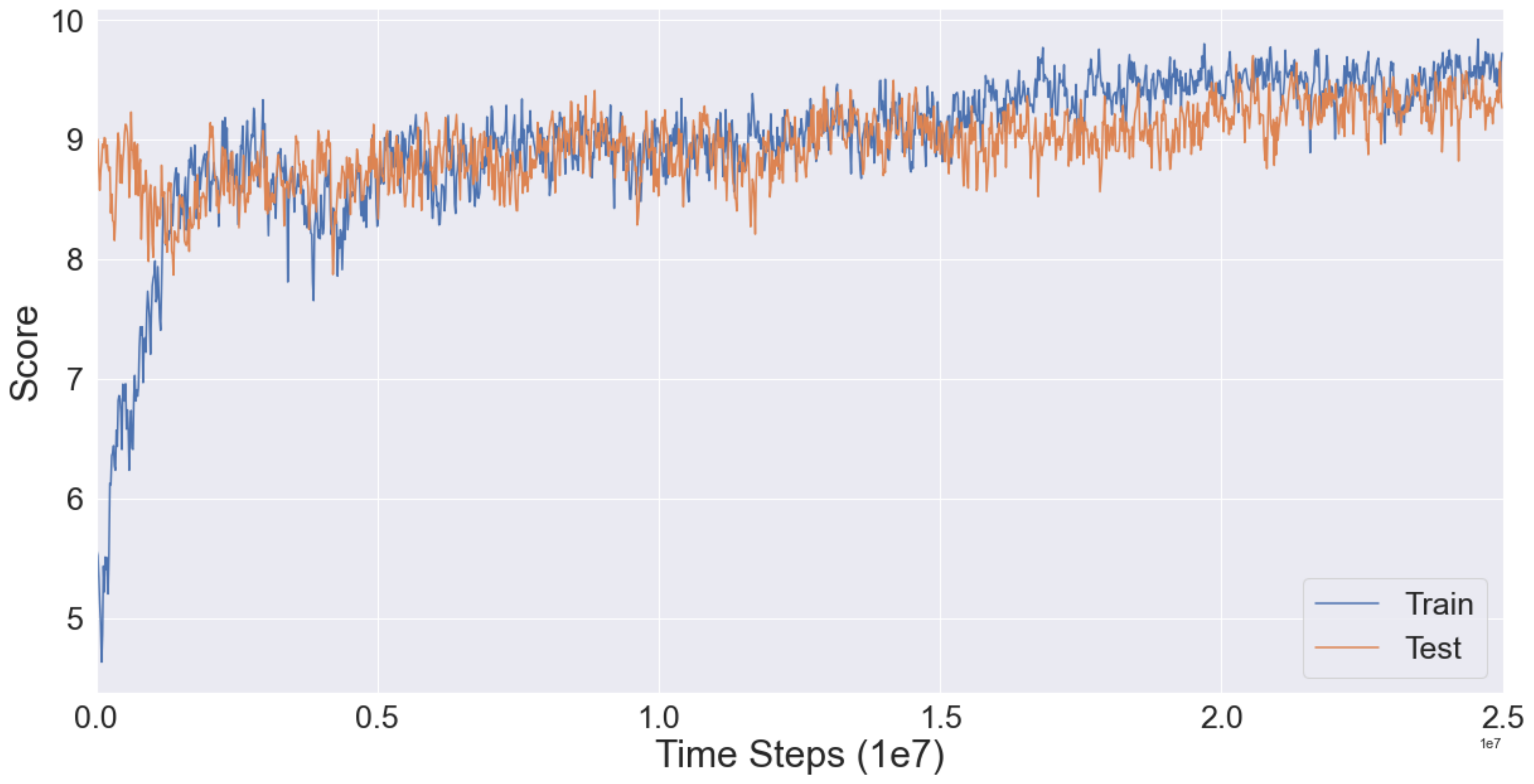}}
\caption{CoinRun - Baseline PPO: Train vs. Test Score}
\label{coinrun_baseline}
\end{center}
\end{figure}

\textbf{CoinRun} In CoinRun, where the agent avoids monsters on a platform to acquire a gold coin, we first evaluate the Procgen PPO baseline algorithm. Figure \ref{coinrun_baseline} confirms the finding from \citealt{Procgen} that there is a gap between training and testing performances, especially during the last 1.5M steps, where the mean task reward is 7.78 for the training levels, and 9.37 for the testing levels. However, the gap is smaller compared to the one reported in \citealt{Procgen} considering the difficulty mode is set to easy.

\begin{figure}[ht]
\begin{center}
\centerline{\includegraphics[width=\columnwidth]{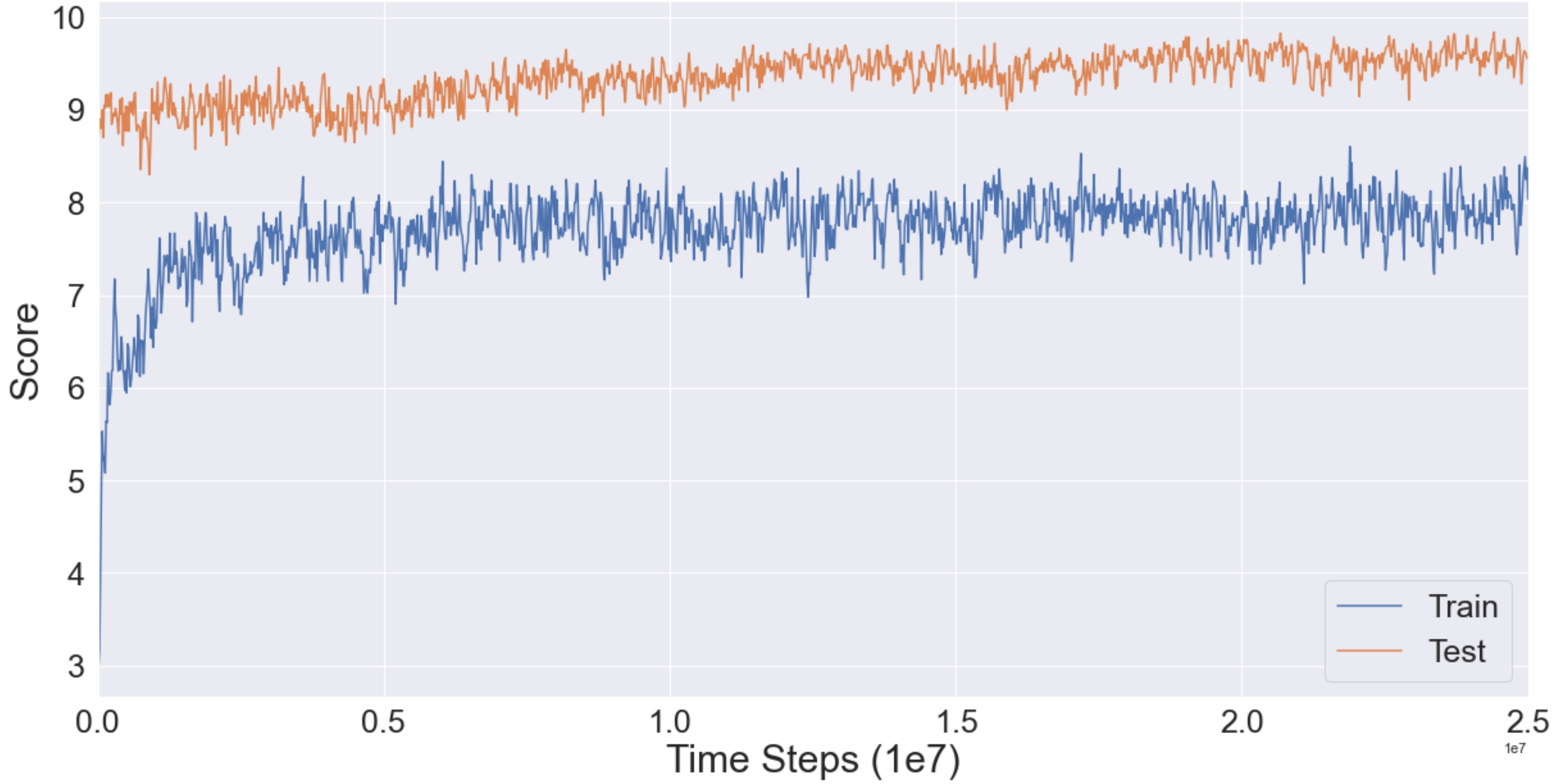}}
\caption{CoinRun - PPO + Normal: Train vs. Test Score}
\label{coinrun_normal}
\end{center}
\end{figure}

The performance of the agent is investigated when it is trained on a combination of both the surprise minimizing reward and the raw task reward from the game. SM reward is defined as the log probability of observed states from normally distributed density model as mentioned in Section 3. The hyper-parameter $\alpha$ of $10^{-4}$ is selected to downscale the SM reward to a similar level as the task reward. In Figure \ref{compare_coinrun_train}, we can see that scores achieved with the combined reward (orange) during 25M steps of training is lower than the baseline (blue). However, on the test set, the agent trained on combined task and Normal SM rewards has comparable scores to the baseline trained on the task reward alone (Figure~\ref{compare_coinrun_test}). The comparison in Figure~\ref{coinrun_normal} shows that with the combined reward, the task rewards on the test set outperforms the training set at all steps. This suggests a simple Gaussian density model can provide additional signal for robustness by maintaining homeostasis.

\begin{figure}[ht]
\begin{center}
\centerline{\includegraphics[width=\columnwidth]{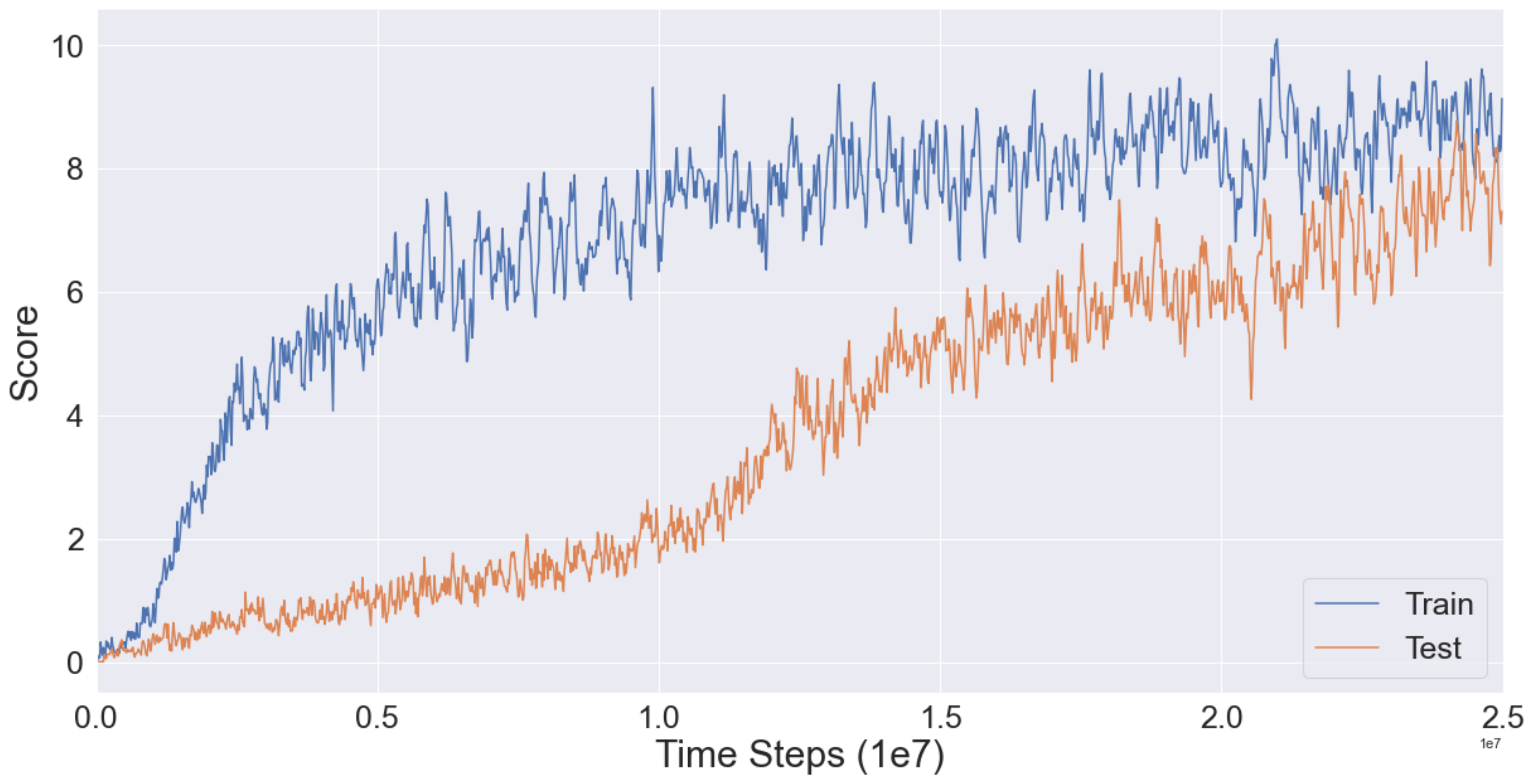}}
\caption{Bossfight - Baseline PPO: Train vs. Test Score}
\label{bossfight_baseline}
\end{center}
\end{figure}

\begin{figure}[ht]
\begin{center}
\centerline{\includegraphics[width=\columnwidth]{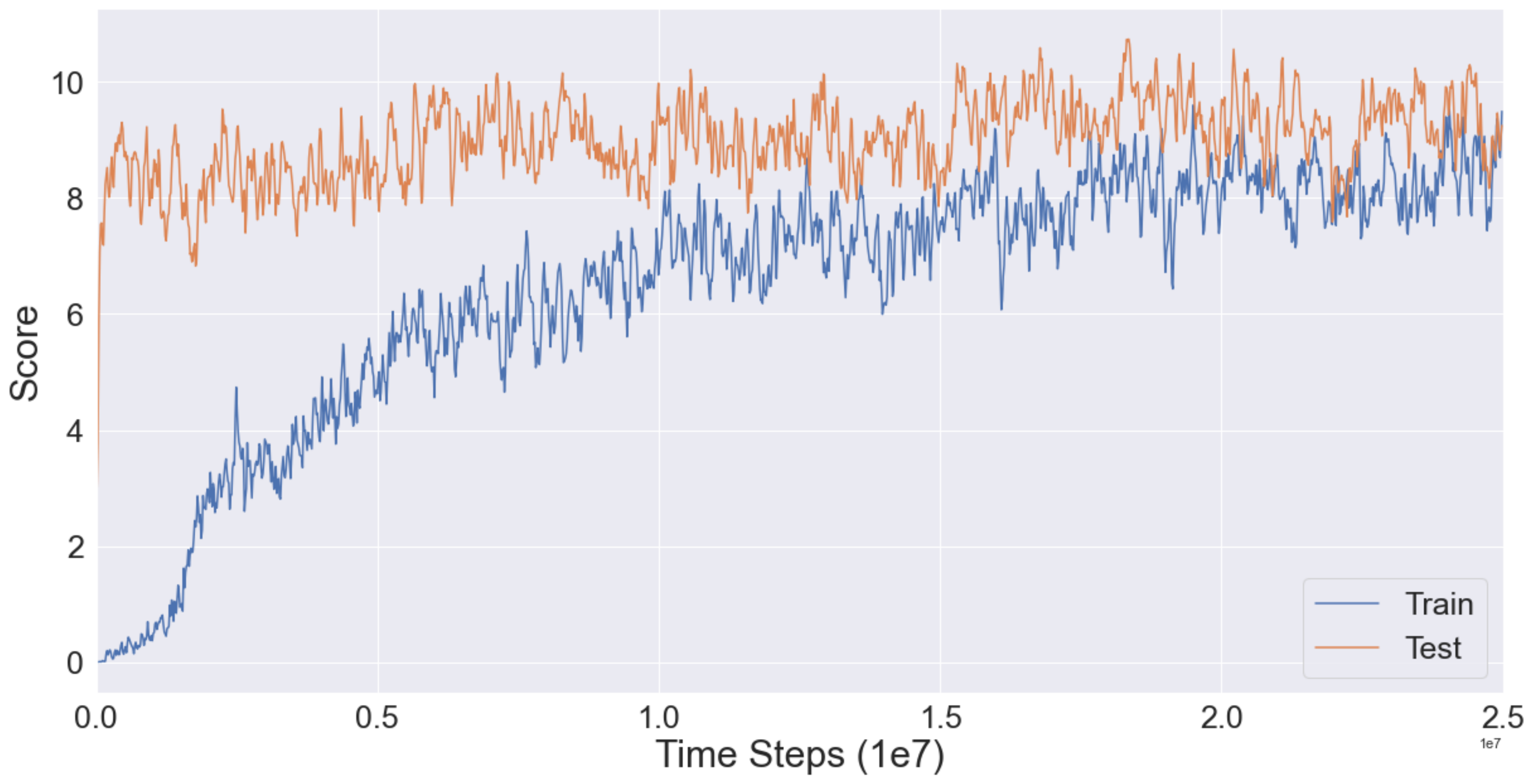}}
\caption{Bossfight - PPO + Normal: Train vs. Test Score}
\label{bossfight_normal}
\end{center}
\vskip -0.2in
\end{figure}

\textbf{BossFight} Similar to CoinRun, same experiments are produced on BossFight. This game is has a more static visual background. The objective is to remain intact from lasers and destroy the boss star ship. In CoinRun, as the agent is centred at all times, the visual background shift according to how the agent moves. In bossfight, only moving parts in BossFight are the boss, the agent and their lasers. An $\alpha$ of $10^{-6}$ works well to downscale SM rewards. It is found that there is a more prominent gap between train and test curves (Figure~\ref{bossfight_baseline}) in the PPO baseline. Furthermore, the learned policy is slow at attaining decent results on the test set, implying a worse generalization performance than CoinRun. By adding the Normal SM reward, we achieve better task rewards on the test levels (Figure~\ref{bossfight_normal} in orange). Comparing test curves (Figure~\ref{compare_bossfight_test}), the addition of a NormalSM reward shows a higher task reward throughout testing, indicating its benefit on generalization.

\subsection{Variational Autoencoder}

The particular behaviour of a surprise minimizing agent is strongly influenced by the choice of state representation \cite{SMiRL}. We further implement a variational autoencoder \cite{VAE, SBVAE} during training to see if a more complex density model can better improve generalization than the simpler Gaussian case. The VAE has convolutional layers similar to \cite{drive}, where they utilized it for autonomous driving environments. The autoencoder has a latent dimension of 100. We train this VAE online to produce latent representations over all the states in the batched observations $\mathbf{s}_t$ to produce mean and variance of a multivariate normal distribution with the same dimension as the latent encoding. Every update to the VAE uses the same batch of observations as the policy update. We optimize the VAE across batched observations to compute the surprise minimizing reward as described in Section 3. Note that due to computational constraints, we train the VAE along with our RL algorithm to 9.3 million steps for both games.

\begin{figure}[ht]
\begin{center}
\centerline{\includegraphics[width=\columnwidth]{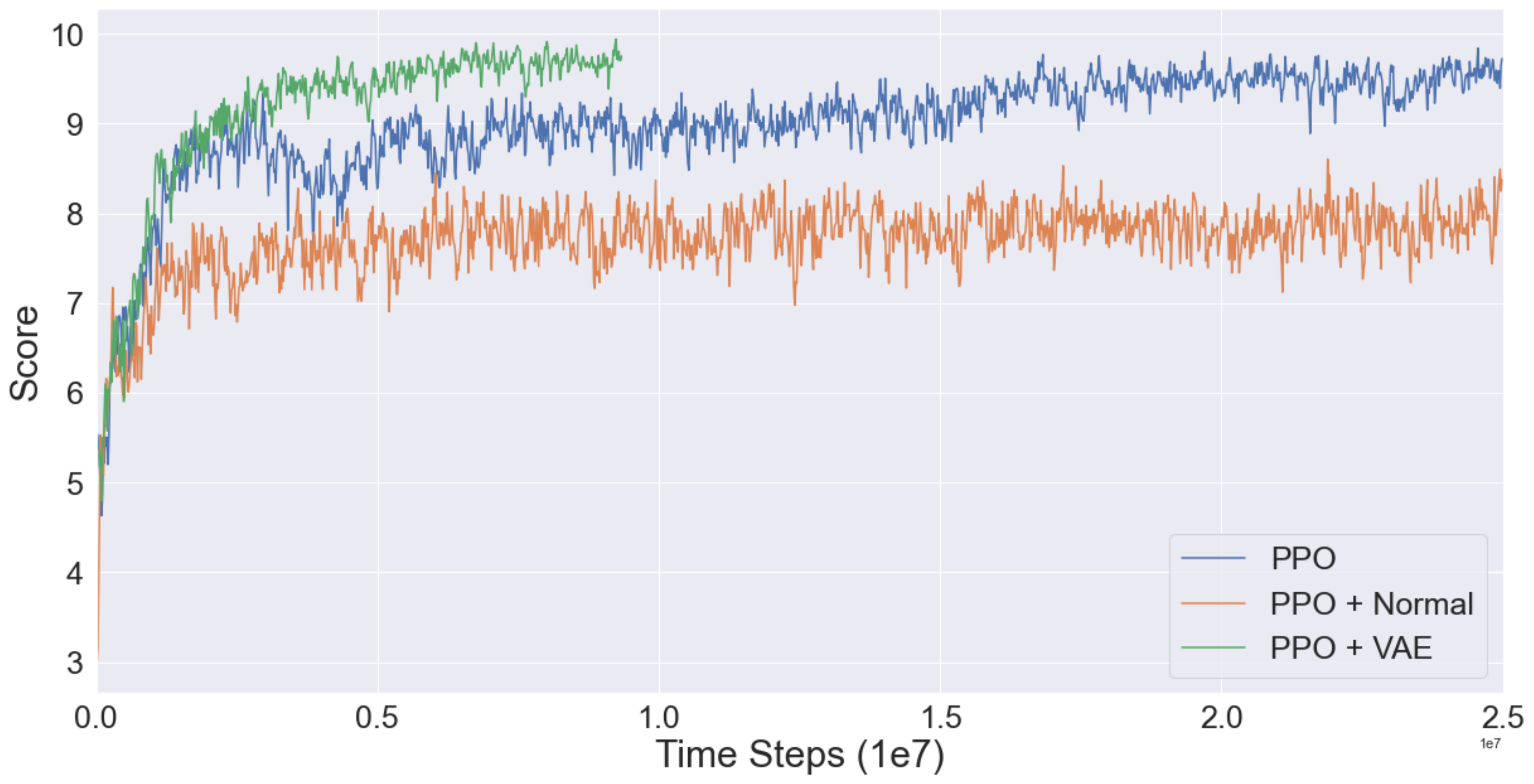}}
\caption{CoinRun Train Scores}
\label{compare_coinrun_train}
\end{center}
\end{figure}

\begin{figure}[ht]
\begin{center}
\centerline{\includegraphics[width=\columnwidth]{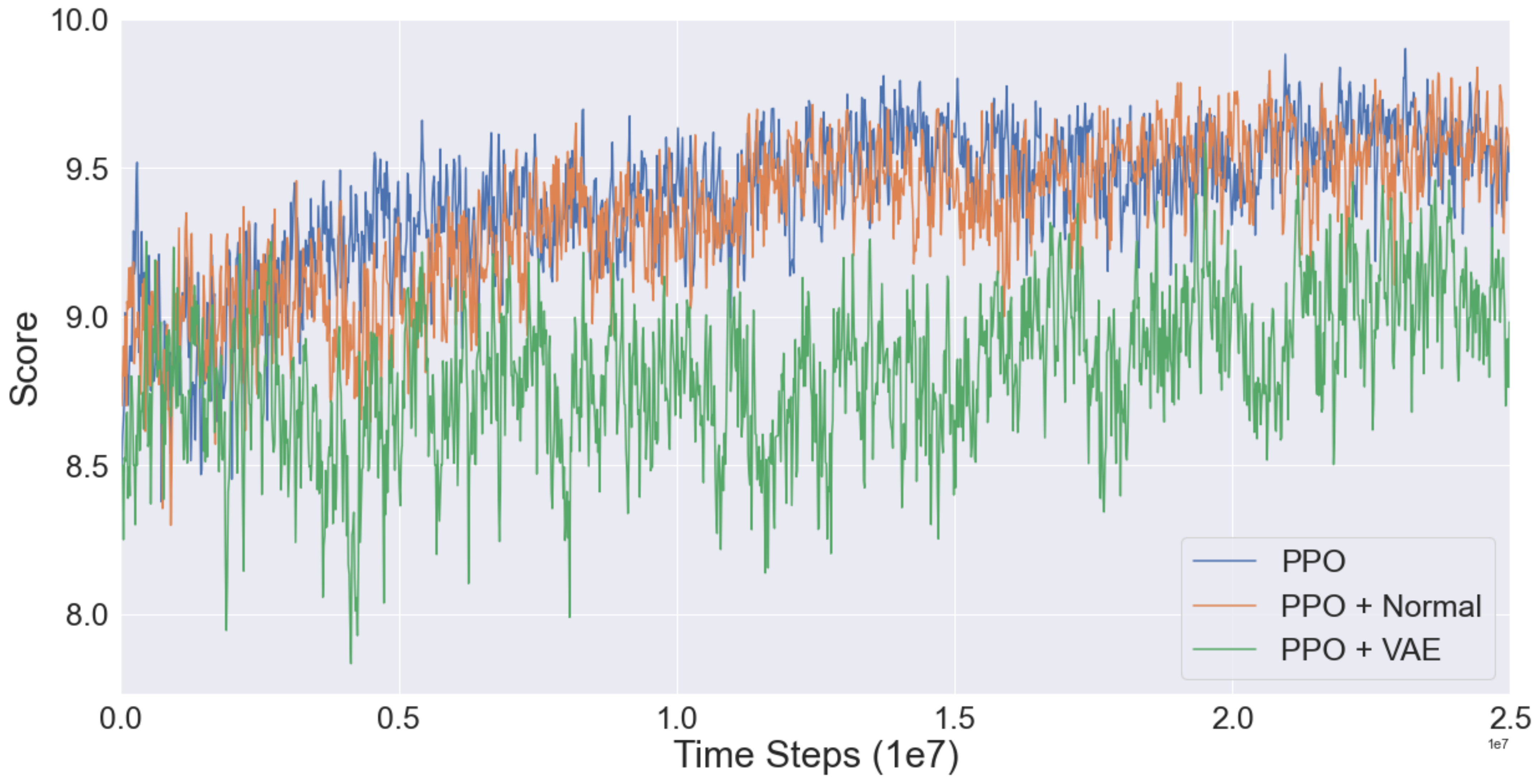}}
\caption{CoinRun Test Scores}
\label{compare_coinrun_test}
\end{center}
\end{figure}

\textbf{CoinRun} We choose $\alpha$ to be $10^{-3}$ for CoinRun. Even though it is stopped early, the task reward for PPO + VAE in Figure~\ref{compare_coinrun_train} (PPO + VAE in green) has already eclipsed the two other methods. SM reward computed from the VAE helps the agent to achieve high task reward significantly faster than the baseline PPO. However, the test results using the policy trained for 9M steps does not achieve better generalization results in Figure~\ref{compare_coinrun_test}. We think that the policy in combination with the additional reward from VAE may overfit to the training levels due to its richer representation power and thus the policy cannot generalize as well to the test levels.

\begin{figure}[ht]
\begin{center}
\centerline{\includegraphics[width=\columnwidth]{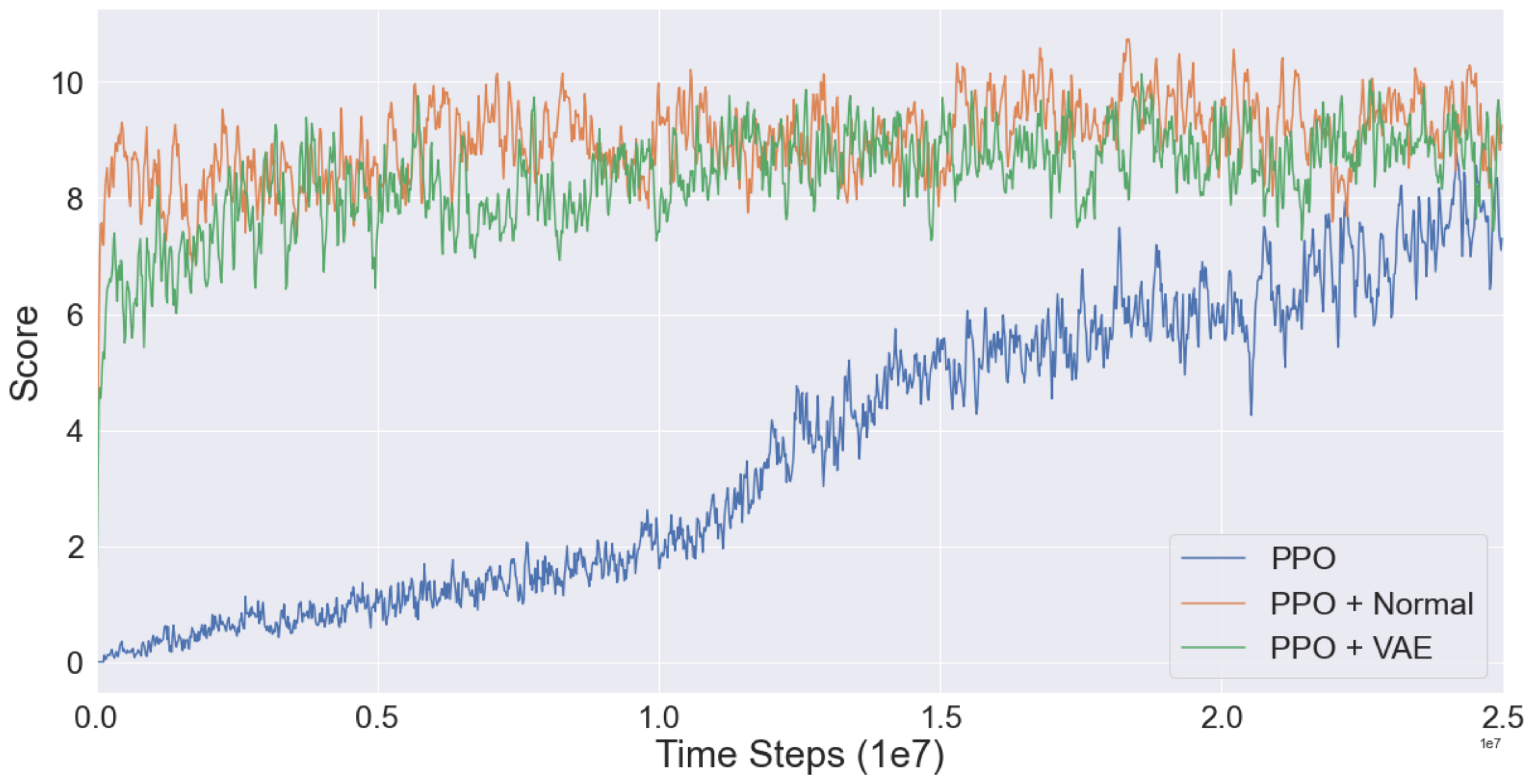}}
\caption{BossFight Test Scores}
\label{compare_bossfight_test}
\end{center}
\end{figure}

\textbf{BossFight} For BossFight, $\alpha$ is chosen to be $10^{-5}$ for balancing the two rewards. In the beginning of testing (see Figure~\ref{compare_bossfight_test}), it can be observed the same possible overfitting effect of the VAE approach, which results in lower test scores. But as testing progresses, rewards get on par with the PPO + Normal method.

\begin{figure}[ht]
\begin{center}
\centerline{\includegraphics[width=\columnwidth]{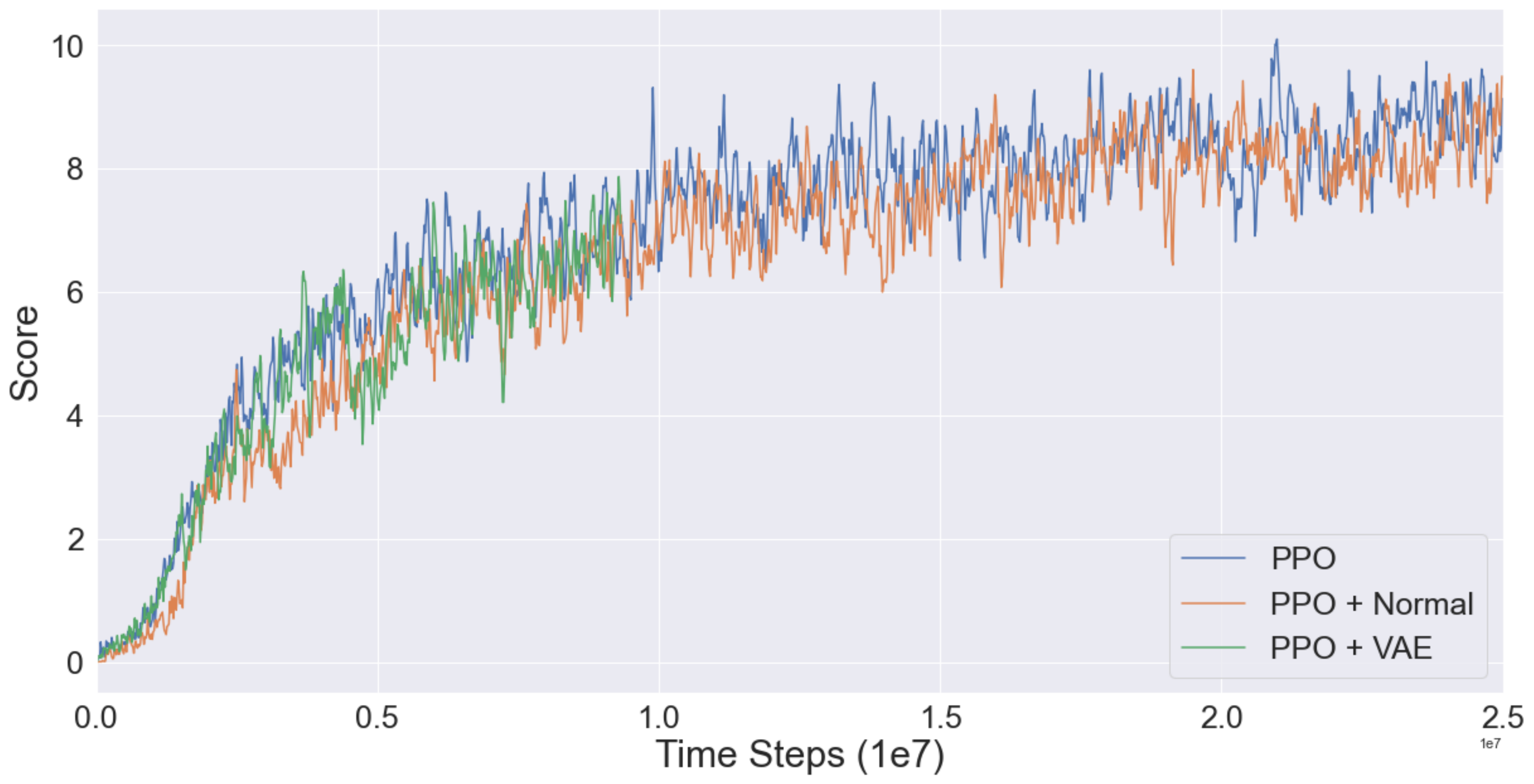}}
\caption{BossFight Train Scores}
\label{compare_bossfight_train}
\end{center}
\end{figure}

\section{Discussions}
As observed, adding surprise minimizing reward can be helpful for generalization in the context of procedurally generated environments. Surprise minimization aims to predict states by learning a density model through historical trajectories under the policy to minimize future surprises. The assumption underlying this framework is that the environment is constantly changing and the agent needs to maintain stability in order to achieve good performance and avoid surprises. This is a suitable framework for robustness since we would like the agent to be able to generalize under entropic environments, where observations are randomized but the task remains the same. The Procgen benchmark supplies sufficient novel stimulus in the training set by varying layout design, background visuals, locations of entities, and so on. By adding the surprise minimizing reward signal, which is learned from stochastic observations, the agent is encouraged toward stability and predictability that counteracts the prevailing entropy in the environment, thus acquiring more robust skills for future surprises that are present in the test set. Additionally, stochasticity in the Procgen environment is injected through the additional SM reward, which might also encourage better generalization performance. 

There are aspects for improvements and further work. First, our analysis is limited to two games in the benchmark and experiments on other games can help better understand the effectiveness of the two methods. As we have discovered, the additional reward works better in BossFight where the agent tries to avoid lasers and remain intact, this is a different objective than searching for a coin in CoinRun. It is unclear how to properly evaluate how much entropy an environment can provide throughout training and which game environments are more or less entropic. If the degree of how entropic a game can be controlled, experiments can be done to understand the extent to which surprise minimizing reward can help. The choice of the density model can also influence the generalization gap. To narrow the gap and prevent overfitting, careful choice of its structure might be needed for a rich model like the variational autoencoder. Further, the choice of the hyperparameter in determining the magnitude of surprise minimizing reward can make or break the generalization performance. The particular values chosen in the paper might not be optimal afterall. During the process of experimentation, the performance is found to become poor when the surprise minimization reward dominates. So how to balance the SM reward properly specific for the task or the density model can be another topic.

\section*{Acknowledgements}
The author thanks Ning Ye for help and discussions, Animesh Garg for the robotics seminar, and the reviewers for helpful commentaries.

\bibliography{reference}
\bibliographystyle{icml2020}


\end{document}